\pdfoutput=1

\documentclass[11pt]{article}

\usepackage{latex/acl}

\usepackage{times}
\usepackage{latexsym}

\usepackage[T1]{fontenc}

\usepackage[utf8]{inputenc}

\usepackage{microtype}

\usepackage{inconsolata}
\usepackage{booktabs}
\usepackage{graphicx}
\usepackage{graphics}
\usepackage{amsmath}
\usepackage{multirow}
\usepackage{amsmath}

\usepackage[normalem]{ulem}

\newcommand{\com}[1]{}

\newcommand{\resolved}[1]{}

\usepackage{amsfonts}
\newcommand{\fw}[0]{ICL on a Budget}

\title{In-Context Learning on a Budget:\\ A Case Study in Token Classification}

\author{Uri Berger$^{1,2}$, Tal Baumel$^3$, Gabriel Stanovsky$^1$ \\
$^1$School of Computer Science and Engineering, The Hebrew University of Jerusalem \\
$^2$School of Computing and Information Systems, University of Melbourne \\
$^3$Microsoft Corporation \\
\texttt{\{uri.berger2, gabriel.stanovsky\}@mail.huji.ac.il} \\
\texttt{tal.baumel@microsoft.com}}

\begin{document}
\maketitle
\begin{abstract}
Few shot in-context learning (ICL) typically assumes access to large annotated training sets. However, in many real world scenarios, such as domain adaptation, there is only a limited budget to annotate a small number of samples, with the goal of maximizing downstream performance.
We study various methods for selecting samples to annotate within a predefined budget, focusing on token classification tasks, which are expensive to annotate and are relatively less studied in ICL setups.
Across various tasks, models, and datasets, we observe that no method significantly outperforms the others, with most yielding similar results, including random sample selection for annotation. Moreover, we demonstrate that a relatively small annotated sample pool can achieve performance comparable to using the entire training set.
We hope that future work adopts our realistic paradigm which takes annotation budget into account.
\end{abstract}

\section{Introduction}
In-context learning (ICL) has emerged as a highly efficient and robust method for various textual tasks.
In this paradigm, a large language model (LLM) is exposed to a small number of annotated samples, termed \emph{demonstration examples}, which are provided as part of the prompt, before the sample which the model is required to annotate, which we will refer to as \emph{inference sample} henceforth. While the reasons for ICL's success are still contested~\cite{min-etal-2022-rethinking, liu-etal-2022-makes}, it has been observed that ICL prompts commonly outperform zero-shot prompts, where no demonstration examples are provided~\cite{brown2020language}.

Furthermore, a recent line of work has found that the choice of demonstration examples can lead to improved results over random demonstration selection. For example, \citet{liu-etal-2022-makes} found that choosing the nearest neighbors of the inference sample in the training set leads to improvements over random demonstration selection on 6 tasks, such as sentiment analysis or question answering. In all of these, the demonstration examples are chosen from large annotated training sets, ranging from 3.5K samples up to 78K samples.  

In this work, we address the following research question: \emph{How can we maximize ICL performance on a given annotation budget?} This question is particularly relevant for real-world domain adaptation settings, where a large pool of annotated samples is unavailable for selecting demonstration examples. Instead, there are large sets of unannotated samples (e.g., raw text in the target domain), and a fixed budget to annotate only a small portion of them.  
As depicted in Figure~\ref{fig:budget}, we define the task as \emph{pool selection}, i.e., selecting a small pool of $k$ examples out of a large corpus of raw texts. These samples are annotated and serve as the available pool for demonstration examples.

\definecolor{my_purple}{rgb}{0.5843, 0.098, 0.9843}

\begin{figure*}[tb!]
    \centering
    \includegraphics[width=\textwidth]{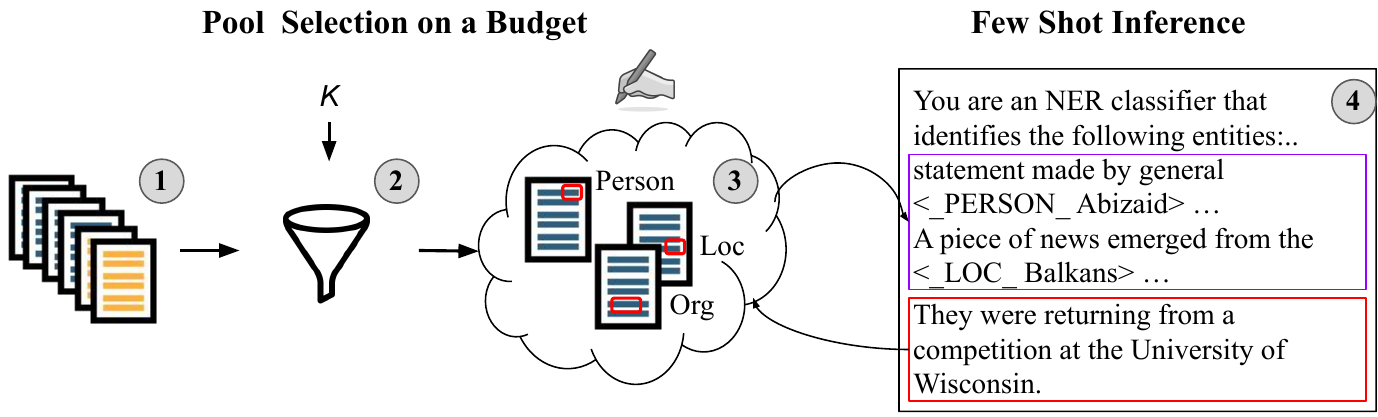}
    \caption{
    Our proposed approach for ICL on a budget, illustrated in four steps: (1) we assume a large pool of \emph{raw} \textcolor{blue}{train} and potentially \textcolor{orange}{test} texts in the target domain; (2) a \emph{pool selection strategy} chooses a subset of $k$ train texts to maximize downstream ICL performance; 
    (3) annotations are collected for the selected pool; 
    (4) an inference prompt is instantiated by choosing the \textcolor{my_purple}{nearest examples} in the pool to the \textcolor{red}{inference sample}. We focus on step (2), experimenting with various pool selection strategies.
\label{fig:budget}}
    
\end{figure*}

We implement several methods for pool selection, e.g., clustering the train set and selecting a representative example from each cluster, and
test them on three token classification tasks: named entity recognition, dependency parsing, and part-of-speech tagging. We select these tasks as they are relatively understudied in the context of ICL and are expensive to annotate due to the need for linguistic expertise and domain-specific knowledge~\cite{chen2015study, zhang-etal-2017-dependency-parsing}. 

We evaluate several state-of-the-art LLMs on token classification benchmarks.
We observe that none of the methods consistently outperforms the others, and, surprisingly, \emph{randomly selecting} samples for the annotation pool performs comparably to more carefully designed approaches in certain scenarios. Furthermore, we find that a relatively small pool ($\sim$200 samples) allows LLMs to perform over 88\% as when demonstrations are selected from the full training set.

We hope that our paradigm is adopted in future work in order to report more realistic ICL performance, and explore new methods for sample selection for other tasks and domains.

\section{\fw{}}
\label{sec:budget}
Here we propose a conceptual framework for ICL in a realistic domain adaptation setting (depicted in Figure~\ref{fig:budget}), where there are no apriori annotated datasets for the target task.
Instead, we assume that there is a large corpus of raw texts in the target domain, and a fixed budget for annotating a small portion of them, such that they
can serve as potential demonstration examples during inference.
Intuitively, the goal of the annotation process is to maximize downstream ICL performance.

Below we formalize the task of pool selection, and describe 4 selection strategies, which aim to maximize different aspects, e.g., coverage of the training set versus coverage of the test set. In the following section we evaluate these approaches for token classification tasks.

\subsection{Pool Selection: Task Definition}
Formally, a \emph{pool selection strategy} is a function:
\begin{equation}
  \label{eq:def}
S_k:\mathcal{P} (\mathcal{D}) \mapsto \mathcal{D}^k
\end{equation}
Where $\mathcal{D}$ represents an unannotated distribution (e.g., all texts in a certain domain), $\mathcal{P} (\mathcal{D})$ is the power set of $\mathcal{D}$, and $k \in \mathbb{N_+}$ is the annotation budget, i.e., the number of samples to annotate. Intuitively, $S$ maps raw sample sets to $k$ train samples (the pool), which are then annotated. In all that follows we denote the input samples set by $D \in \mathcal{P} (\mathcal{D})$. Typically, $k << |D|$, indicating that the annotation budget for a new domain can only annotate a small portion of its available texts.

Furthermore, we assume a similarity function $\phi$:
\begin{equation}
\label{eq:sim}
\phi : \big(\mathcal{D} \times \mathcal{D}\big) \mapsto \mathbb{R}
\end{equation}
In the scope of this work (similar to previous work~\citep{liu-etal-2022-makes}) samples are embedded into $\mathbb{R}^m$, where $m$ is the text embedding dimension, and $\phi$ is the cosine similarity of these vectors. In particular, we use a sentence transformer~\citep{reimers-2019-sentence-bert} trained over MPNet~\citep{song2020mpnet}.



\subsection{Pool Selection Strategies} \label{sec:pool_selection}
Below we describe 4 pool selection strategies which follow the definition in Equation~\ref{eq:def}.





\paragraph{Central.}
Select the $k$ samples from $D$ that are closest to the Euclidean center of $D$.


\paragraph{Cluster.}
Cluster $D$ into $k$ clusters and for each cluster center choose the most similar sample from $D$.
This method aims to maximize the coverage of the expected training distribution.
This strategy was proposed in \citet{chang-etal-2021-training} for selecting examples for fine-tuning.

\paragraph{Vote-k~\citep{su2022selective}.}
Selects a $k$-sized subset of $D$ such that the samples are diverse and dissimilar from one another, through a two-step process which uses the LLM's confidence to bucket the different samples. See \citet{su2022selective} for more details.

\paragraph{Random.}
Randomly select $k$ samples from $D$.

\section{Evaluation}
\begin{figure*}[tb!]
    \centering
    \includegraphics[width=\textwidth]{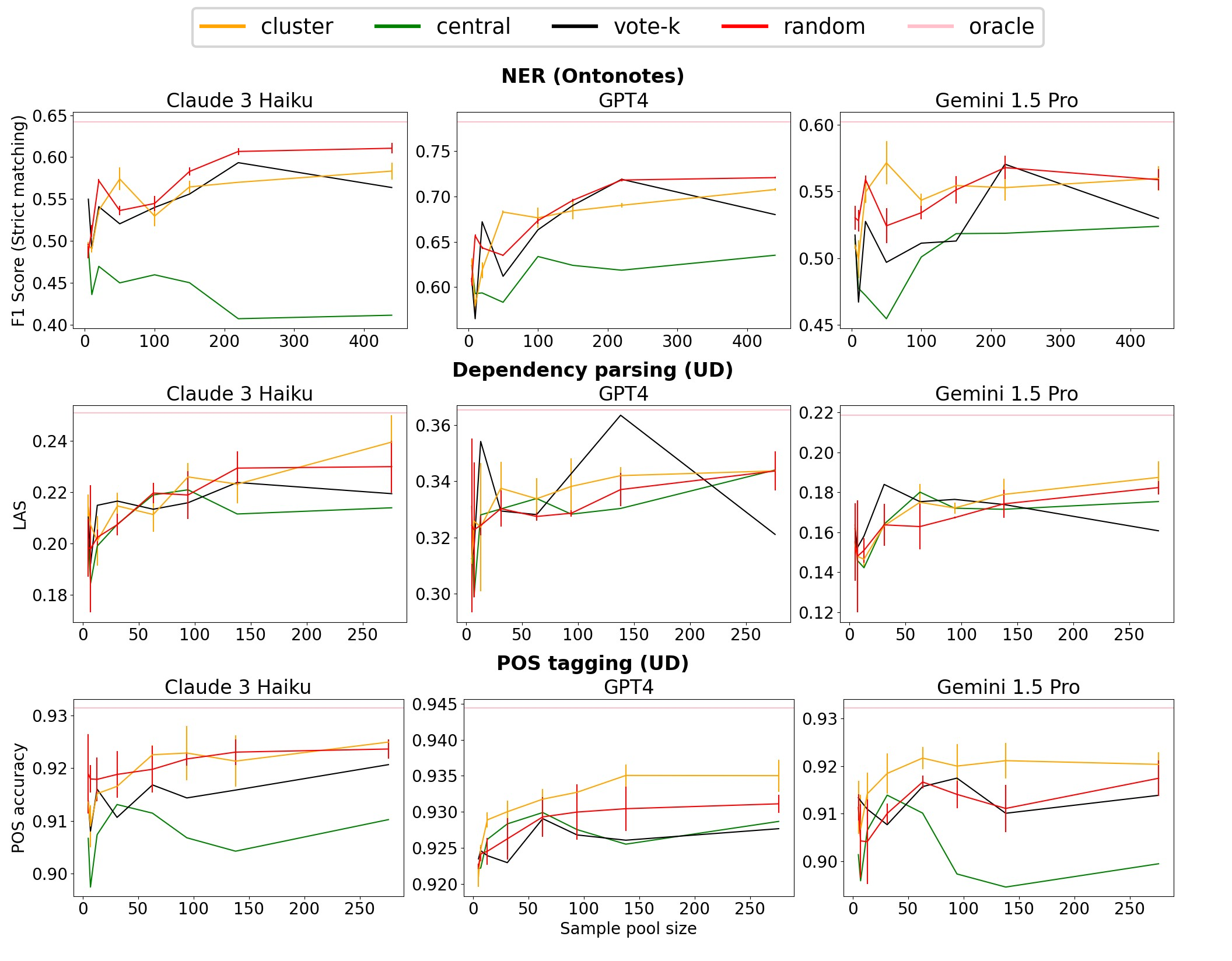}
    \caption{Results for NER (top), dependency parsing (center), and POS tagging (bottom) using Claude 3 Haiku (left), GPT4 (middle) and Gemini 1.5 Pro (right), as a function of the size of sample pool. In methods with a random component we run 3 trials and plot error bars showing the standard deviation. Oracle: using the full training set as the pool.}
    \label{fig:all_results}
\end{figure*}

We focus on token classification tasks, as they are both understudied in the context of ICL, while traditionally expensive to annotate at scale.
Importantly, as is commonly done in active learning scenarios~\citep[e.g.,][]{shen2017deep, liu2020ltp}, we use existing datasets to \emph{simulate} the process shown in Figure~\ref{fig:budget}, where instead of collecting manual annotations, we sample from the existing annotations.

\subsection{Experimental Setup} \label{sec:experimental_setup}

\paragraph{Tasks, datasets, and models.}
We test three tasks using  two English datasets: Ontonotes 5.0 for NER~\cite{hovy-etal-2006-ontonotes}, and the Universal Dependencies  corpus for dependency parsing and POS tagging (UD; \citealp{nivre-etal-2016-universal}).\footnote{
We use the HuggingFace versions for both datasets with the \emph{english\_v12} configuration for \href{https://huggingface.co/datasets/conll2012_ontonotesv5}{Ontonotes} and the \emph{en\_ewt} configuration for \href{https://huggingface.co/datasets/universal-dependencies/universal_dependencies}{UD}.} Both datasets contain texts from various domains, including news, conversational, weblogs, web forums, and more.
Due to budget constrains we limit the size of each test set by randomly sampling 1000 samples.  
We experiment with a diverse set of 9 different LLMs, ranging from open to closed models in various parameter sizes: Llama-2~\cite{touvron2023llama}, Mistral~\cite{jiang2023mistral}, Starling~\cite{starling2023}, Vicuna~\cite{vicuna2023}, Mixtral~\cite{jiang2024mixtral}, phi-2~\cite{javaheripi2023phi}, GPT-4~\cite{achiam2023gpt}, Claude 3 Haiku,\footnote{\href{https://www.anthropic.com/news/claude-3-family}{www.anthropic.com/news/claude-3-family}} and Gemini 1.5 Pro~\cite{reid2024gemini}.


\paragraph{Prompt and evaluation metric.}
For each task, we prompt the models by describing the task and the expected output format,
followed by 5 demonstration examples and the current inference sample.
Since POS tagging serves as a precursor to dependency parsing, both tasks are handled using a shared prompt.
Following \citet{zhao2021calibrate},
we sort the examples such that the most similar example is the last one, based on the similarity function $\phi$ described in Equation~\ref{eq:sim}.
See Appendix~\ref{sec:app_prompt_template} for the full prompt templates.

For NER evaulation we use \emph{strict} matching~\cite{semeval2013}, where a predicted entity is considered correct if it matches both exact span boundaries and entity type. For dependency parsing we use \emph{labeled attachment score (LAS)}, which measures the accuracy of both the correct head and dependency label for each token. For POS tagging, we use the \emph{POS accuracy}.

\paragraph{Pool sizes.}
We experiment with sample pools of size 0.1\% - 10\% of the maximum pool size, defined as the number of unique samples used when considering the full training set as the sample pool.

\paragraph{Oracle.}
As a reference, we
report the performance of using the full training set for each dataset and model, compared to our budget-constrained approach.

\subsection{Results} \label{sec:results}

Results for the three best-performing models (Claude 3 Haiku, GPT-4, and Gemini 1.5 Pro) on all tasks are presented in Figure~\ref{fig:all_results}.  The other smaller models we test
are not able to produce outputs of the requested format in more than 50\% of the cases, and hence cannot be meaningfully compared against these models, which adhere to the correct format in roughly 97\% of the cases.
We now discuss findings reflected in these results and conduct further analysis. 

\paragraph{The choice of few-shot examples matter in token classification tasks.}
We observe a large variation in performance when selecting different demonstration examples in all configurations. While this was observed in other tasks~\citep{zhang-etal-2022-active}, to the best of our knowledge, this is the first time this was shown for token classification tasks.

\paragraph{Most pool selection methods perform similarly, random is surprisingly good.}
None of the four methods consistently outperform the others.
Surprisingly, we note that random performs similarly to other methods.


\paragraph{Very small pool sizes can approximate the full training corpus.}
All sampling methods require a pool of only 220 samples for Ontonotes, and 138 samples for UD, to achieve over 88\% of the oracle performance across all configurations.

\paragraph{ICL on a budget lags behind state-of-the-art.}
We compare the ICL results to state-of-the-art results in Table~\ref{tab:sota}. In NER and dependency parsing, fine-tuned methods vastly outperforms using a limited annotation budget. In POS tagging which is considered an easier task, using 138 samples achieves 96\% of the state-of-the-art.

\begin{table}[t]
\centering
\begin{tabular}{l|lll}
\toprule
 & NER & DP & POS \\
\midrule
Central & 67 & 40 & 96 \\
Cluster & 75 & 41 & 96 \\
Vote-k & 78 & 44 & 96 \\
Random & 78 & 41 & 96 \\
\bottomrule
    \end{tabular}
    \caption{GPT-4 percentage in performance out of the state-of-the-art, when using 5\% of the samples used in the fully labeled train set method.}
    \label{tab:sota}
\end{table}



\section{Related Work}

Similar to pool selection, active learning~\cite{shen-etal-2017-deep} also aims to select samples for annotation rather than assuming all samples are annotated. However, active learning operates during training and relies on access to an oracle or intermediate model results (e.g., confidence scores), whereas pool selection assumes no access to the model during training and only relies on observing the outputs of the model.

Recently, \citet{su2022selective} introduced a pool selection method as a method for improving downstream performance, which we evaluate as one of our approaches for annotation pool selection (\emph{cluster}). 
Our work is conceptually different in that it proposes a realistic paradigm under which to examine ICL performance where there are no annotated samples. Subsequently, we differ from them in that we study three different token level tasks, different pool selection methods, and particularly focus on the effect of the pool size on downstream performance.

\section{Conclusion}
We proposed the framework of ICL on a budget and studied different methods for pool selection, focusing on token classification tasks. We hope this work will inspire more realistic assumptions on the amount of labeled data used in different ICL settings.

\section{Limitations}
We tested our ICL on a budget approach on a single class of tasks (token classification), because it has high annotation cost and it was relatively less studied in the context of ICL. It is possible that other tasks will show different trends, hence we stress that the contribution here is the methodological approach, rather than advocating for one particular sampling strategy. 

\bibliography{anthology,custom}

\appendix

\section{Prompt Template}
\label{sec:app_prompt_template}
In this section we describe the prompt templates we use in our experiments.

\subsection{NER task description}

{\fontfamily{Arial}\selectfont
You are an NER classifier that identifies the following entities: Person \_\_PER\_\_,  Organization \_\_ORG\_\_, Geo-Political \_\_GPE\_\_, Location \_\_LOC\_\_, Facility \_\_FAC\_\_, Work-of-Art \_\_WOA\_\_, Event \_\_EVE\_\_, Product \_\_DUC\_\_, Language \_\_ANG\_\_ use angle brackets to tag in-line, please don't include any additional information other than the annotated sentence and keep original spacing.
}

\begin{figure*}[tb]
    \centering
    \includegraphics[width=\textwidth]{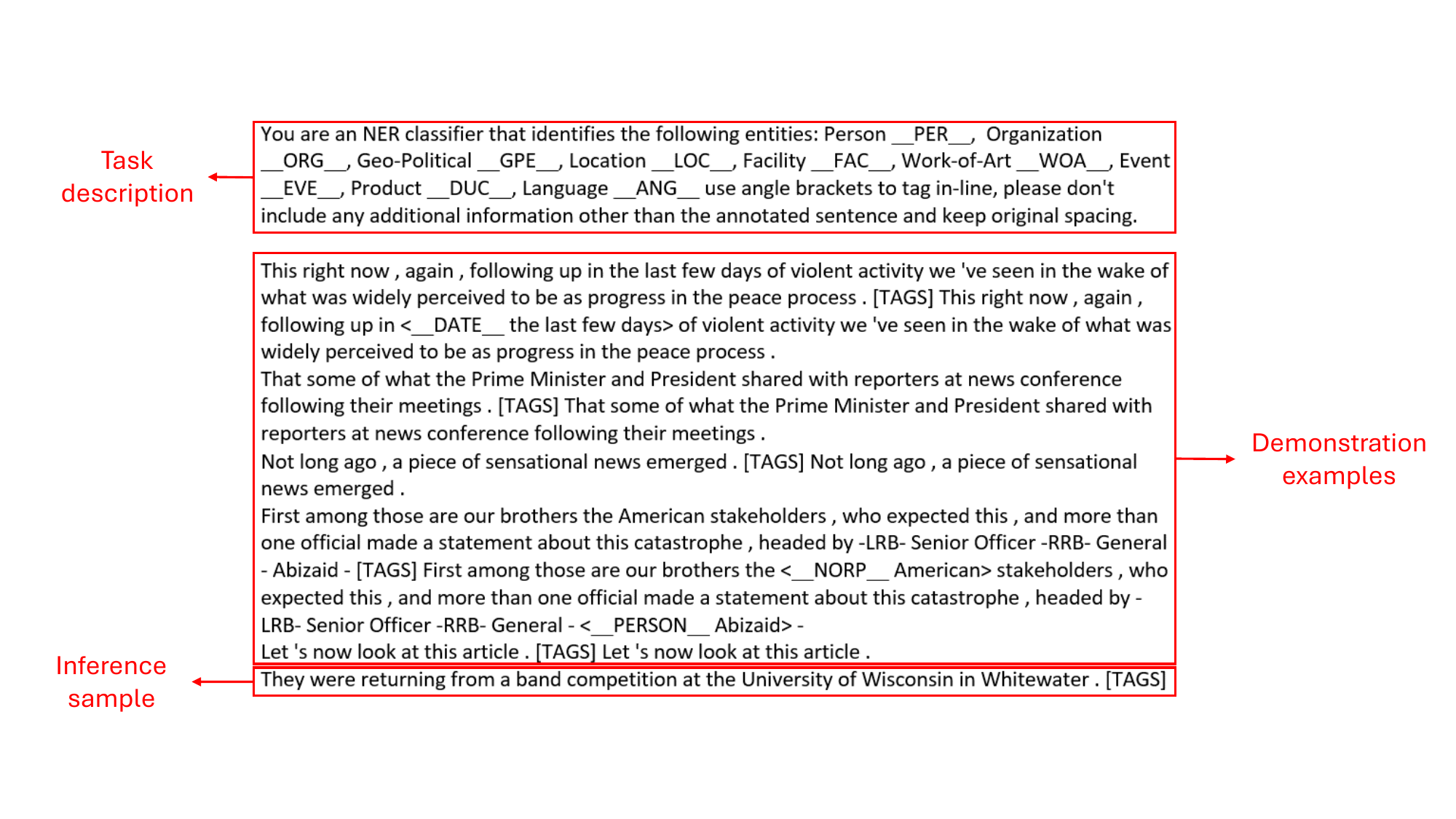}
    \caption{An example NER prompt used in our study.}
    \label{fig:prompt_example}
\end{figure*}

\begin{figure*}[tb]
    \centering
    \includegraphics[width=\textwidth]{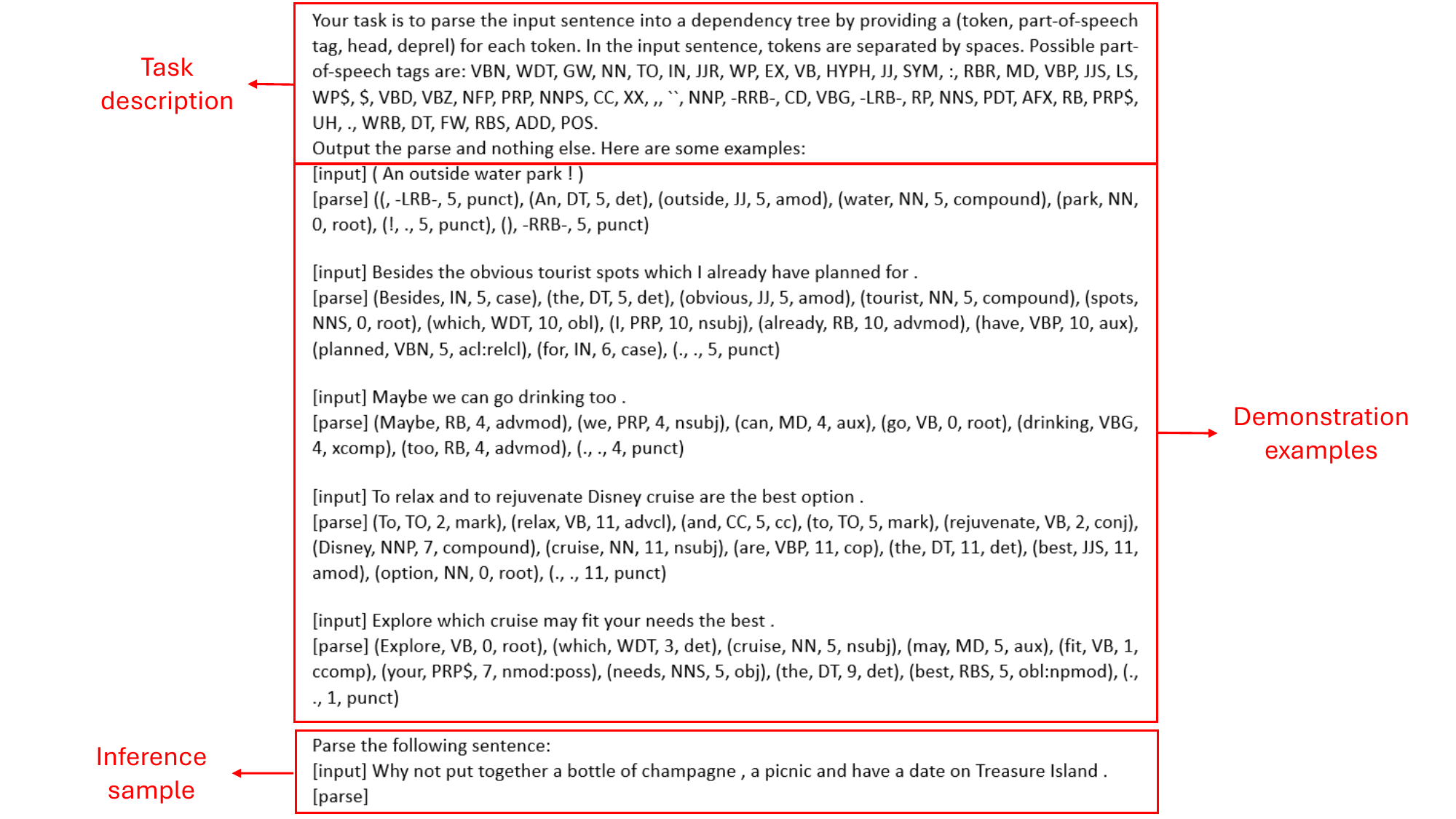}
    \caption{An example dependency parsing prompt used in our study.}
    \label{fig:dep_parse_prompt_example}
\end{figure*}

\subsection{Dependency parsing and POS tagging task description}

{\fontfamily{Arial}\selectfont
Your task is to parse the input sentence into a dependency tree by providing a (token, part-of-speech tag, head, deprel) for each token. In the input sentence, tokens are separated by spaces. Possible part-of-speech tags are:

VBN, WDT, GW, NN, TO, IN, JJR, WP, EX, VB, HYPH, JJ, SYM, :, RBR, MD, VBP, JJS, LS, WP\$, \$, VBD, VBZ, NFP, PRP, NNPS, CC, XX, ,, ``, NNP, -RRB-, CD, VBG, -LRB-, RP, NNS, PDT, AFX, RB, PRP\$, UH, ., WRB, DT, FW, RBS, ADD, POS, ''

Output the parse and nothing else. Here are some examples:
}

\subsection{Prompt design}

We first describe the task in question, as outlined above.
Next, we add the demonstration examples. For GPT and Claude we add the demonstration examples as follows. For each demonstration example, we use LangChain's\footnote{\href{https://www.langchain.com/}{langchain.com}} HumanMessage class for the original sentence, followed by an AIMessage for the annotated sentence. Finally, we add the inference sample as a HumanMessage.

In Gemini, for which these classes are not implemented, we choose a different strategy: for each demonstration example, we add the original sentence followed by the separator token ([TAGS] for NER, [PARSE] for dependency parsing and POS tagging), and then the annotated sentence. Finally, we add the inference sample, followed by the separator token. Figures~\ref{fig:prompt_example} and \ref{fig:dep_parse_prompt_example} demonstrates an NER 
and dependency parsing prompts for Gemini, respectively.


\section{Performance-Diversity Correlation} \label{sec:app_perf_div_corr}
\citet{min-etal-2022-rethinking} study the factors that impact performance in ICL and find that the coverage of the label space by the demonstration examples has a
strong effect on performance.
Drawing inspiration from their findings, we examine whether the diversity in the labels of demonstration examples is correlated with performance.
To this end, for each dataset, pool selection method and pool size, we count how many instances of each label (entity for NER, dependency label for dependency parsing, POS tag for POS tagging) were presented in the demonstration examples in the sample pool, and compute the entropy of these counts as a proxy for diversity. For each model, we then compute the Pearson correlation between these entropy values and the model's scores. Table~\ref{tab:perf_div_corr} presents the correlations. Correlation is high ($>0.5$) for all models in the dependency parsing and POS tagging tasks.

\begin{table}[t]
\centering
\begin{tabular}{l|lll}
\toprule
 & GPT & Claude & Gemini \\
\midrule
NER & 0.54 & 0.36 & 0.46 \\
DP & 0.64 & 0.62 & 0.64 \\
POS & 0.53 & 0.63 & 0.80 \\
\bottomrule
    \end{tabular}
    \caption{Pearson correlation of performance with label diversity in the sample pool, as measured by the entropy of entities. All results are significant.}
    \label{tab:perf_div_corr}
\end{table}


\end{document}